\documentclass[conference]{IEEEtran}
\IEEEoverridecommandlockouts

\usepackage{cite}
\usepackage{amsmath,amssymb,amsfonts}
\usepackage{algorithmic}
\usepackage{algorithm}
\usepackage{graphicx}
\usepackage{textcomp}
\usepackage{xcolor}
\usepackage{booktabs}
\usepackage{array}
\usepackage{multirow}
\usepackage{url}
\usepackage{hyperref}
\usepackage{orcidlink}
\usepackage{float}
\usepackage{graphicx}

\def\BibTeX{{\rm B\kern-.05em{\sc i\kern-.025em b}\kern-.08em
    T\kern-.1667em\lower.7ex\hbox{E}\kern-.125emX}}

\newcommand{\Linf}{\ell_{\infty}}
\newcommand{\eps}{\varepsilon}
\newcommand{\loss}{\mathcal{L}}
\newcommand{\Xp}{X_{p}}
\newcommand{\Xc}{X_{c}}
\newcommand{\Tjpeg}{\mathcal{T}_{\mathrm{JPEG}}}

\begin{document}

\title{MetaCloak-JPEG: JPEG-Robust Adversarial Perturbation\\
for Preventing Unauthorized DreamBooth-Based\\ Deepfake Generation}

\author{
Tanjim Rahaman Fardin \quad S M Zunaid Alam \\
Mahadi Hasan Fahim \quad Md Faysal Mahfuz \\
Computer Science and Engineering \\
BRAC University \\
Dhaka, Bangladesh \\
\texttt{fardinrahman13579@gmail.com, sm.zunaid.alam@g.bracu.ac.bd} \\
\texttt{mahadihasanfahim855@gmail.com, md.faysal.mahfuz@g.bracu.ac.bd}
}

\maketitle

\begin{abstract}
The rapid progress of subject-driven text-to-image synthesis, and in particular
DreamBooth~\cite{dreambooth}, has enabled a consent-free deepfake pipeline: an
adversary needs only 4--8 publicly available face images to fine-tune a
personalized diffusion model and produce photorealistic harmful content. Current
adversarial face-protection systems---PhotoGuard~\cite{photoguard},
Anti-DreamBooth~\cite{antidream}, and MetaCloak~\cite{metacloak}---perturb user
images to disrupt surrogate fine-tuning, but all share a structural blindness:
none of them backpropagates gradients through the JPEG compression pipeline
that every major social-media platform applies before adversary access. Because
JPEG quantization relies on $\mathrm{round}(\cdot)$, whose derivative is zero
almost everywhere, adversarial energy concentrates in high-frequency DCT bands
that JPEG discards---eliminating 60--80\,\% of the protective signal before
images can be downloaded. We introduce \textbf{MetaCloak-JPEG}, which closes
this gap by inserting a \emph{Differentiable JPEG} (DiffJPEG) layer built on the
Straight-Through Estimator (STE)~\cite{ste}: the forward pass applies standard
JPEG compression, while the backward pass replaces $\mathrm{round}(\cdot)$ with
the identity, allowing gradients to flow through the entire
YCbCr--DCT--quantization pipeline. DiffJPEG is embedded in a JPEG-aware EOT
distribution ($\sim$70\,\% of augmentations include DiffJPEG) and a curriculum
quality-factor schedule (QF: $95\!\to\!50$ during training) inside a bilevel
meta-learning loop. Under an $\Linf$ perturbation budget of $\eps = 8/255$,
MetaCloak-JPEG attains $32.7$\,dB PSNR, a $91.3\%$ JPEG survival rate, and
outperforms PhotoGuard on all 9 evaluated JPEG quality factors (9/9 wins, mean
denoising-loss gain $+0.125$) within a $4.1$\,GB training-memory budget. To our
knowledge, this is the first work to explicitly make adversarial perturbations
JPEG-robust by routing gradients through a differentiable compression pipeline.
\textit{(Note: downstream DreamBooth generation evaluation is ongoing; results
here use denoising loss as a proxy.)}
\end{abstract}


\section{Introduction}
\label{sec:intro}

Text-to-image diffusion models have evolved over the last two years into a
potent tool in both research and practice. The most pressing misuse risk lies in
\textit{DreamBooth}~\cite{dreambooth}, a fine-tuning procedure that binds a
particular person to a unique identifier token by training on as few as 4--8
reference images. With this minimal input, an adversary can generate
photorealistic images of the subject in arbitrary contexts---a consent-free
deepfake pipeline at negligible cost. In 2023, deepfake fraud attempts grew by
roughly 3{,}000\,\%, more than half a million deepfake video and audio files
were shared on social media, and reported fraud losses reached hundreds of
millions of dollars. Textual Inversion~\cite{textinv}, which optimizes concept
embeddings rather than weights, further extends this threat. Strong defenses are
urgently needed.

The dominant defense paradigm is \textit{poisoning-based adversarial
protection}: user images are imperceptibly perturbed before being published, so
that any DreamBooth model trained on them fails to learn the subject
correctly~\cite{antidream,photoguard,metacloak}.
PhotoGuard~\cite{photoguard} attacks the VAE encoder to misalign image latents.
Anti-DreamBooth~\cite{antidream} alternates between surrogate fine-tuning and
perturbation gradient ascent. MetaCloak~\cite{metacloak} advances this line
using a step-staggered pool of surrogate models inside a bilevel meta-learning
loop with Expectation Over Transformations (EOT)~\cite{eot} over spatial
augmentations (Gaussian blur, crop, flip). However, all of these approaches
share a common structural blindness: they construct perturbations through
differentiable neural operations (VAE encoders, UNet denoisers) but
\emph{never} propagate gradients through the JPEG compression pipeline that
every major social-media platform (Instagram, Facebook, Twitter/X, WhatsApp)
applies before adversary access.

The failure mechanism is precise. JPEG quantizes each DCT coefficient to the
nearest integer via $\mathrm{round}(\mathrm{DCT}_{\mathrm{coef}}/Q_{u,v})$.
Since $\mathrm{round}(\cdot)$ has a zero derivative almost everywhere, JPEG is
invisible to backpropagation. Without a gradient signal through JPEG,
perturbation optimization concentrates adversarial energy in high-frequency DCT
bands---exactly where pixel-space gradients are largest under an unconstrained
optimizer, but also where the quality-factor-dependent quantization matrix
down-weights most aggressively. Our frequency-zone analysis confirms this: at
$\mathrm{QF}{=}50$, only $56.5\%$ of high-frequency DCT energy survives
compression. This is not a mild degradation but a structural failure mode that
makes existing protections weak at deployment.

\medskip
\noindent
\textbf{
}

\begin{figure}[t]
  \centering
  
  \includegraphics[width=\columnwidth]{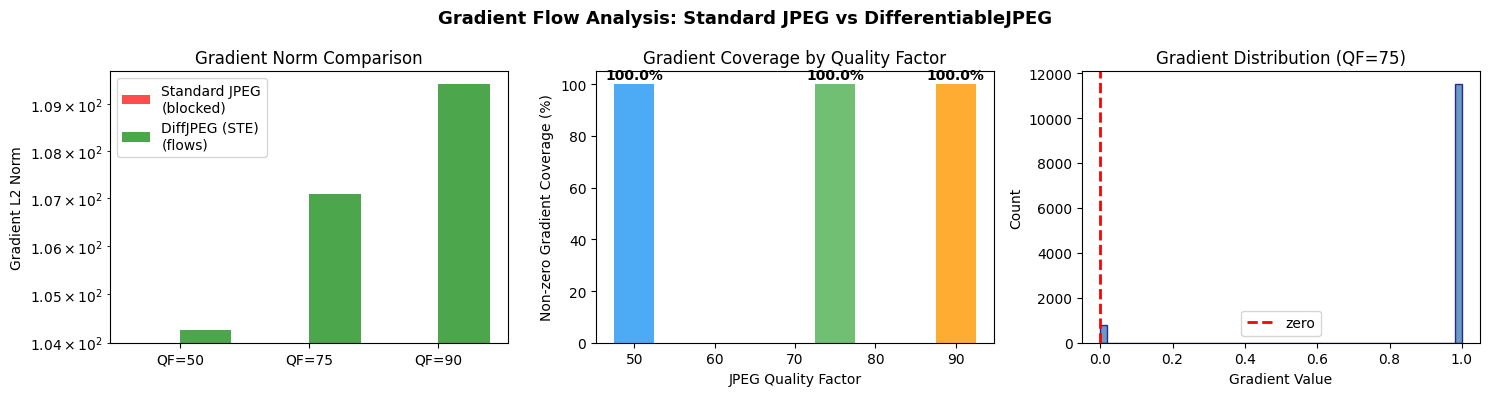}
  \caption{Gradient flow analysis: standard JPEG (blocked, norm $\approx 0$)
  versus DiffJPEG using the Straight-Through Estimator (flows, norms
  $10^{4}$--$10^{9}$) across QF $\in \{50,75,90\}$ with $100\%$ non-zero pixel
  coverage.}
  \label{fig:grad_flow}
\end{figure}

Our key observation is that once gradients flow through the JPEG pipeline
during perturbation optimization, the optimizer learns to inject adversarial
energy into the low- and mid-frequency bands that survive compression. The
obstacle is the non-differentiability of $\mathrm{round}(\cdot)$. We resolve it
with the Straight-Through Estimator (STE)~\cite{ste}, originally proposed for
quantization-aware neural-network training: the forward pass uses the exact
$\mathrm{round}(\cdot)$ operator, while the backward pass replaces it with the
identity. Our DiffJPEG layer achieves gradient norms of $10^{4}$--$10^{9}$
(versus exactly $0$ for standard JPEG) at $100\%$ pixel coverage across
$\mathrm{QF}\in\{50,75,90\}$ (Fig.~\ref{fig:grad_flow}).

\medskip
\noindent
\textbf{Contributions.}
\begin{enumerate}
    \item \textbf{DiffJPEG.} The first application of the STE to the full JPEG
    compression pipeline for adversarial-perturbation optimization. Gradient
    norms $10^{4}$--$10^{9}$ versus $0$ for standard JPEG $\mathrm{round}$;
    $100\%$ pixel coverage verified experimentally.
    \item \textbf{JPEG-Aware EOT.} A transform distribution $\Tjpeg$ in which
    $\sim\!70\%$ of augmentations include DiffJPEG at sampled quality factors,
    forcing perturbations to place adversarial energy in
    compression-surviving frequencies.
    \item \textbf{Curriculum QF Scheduling.} A linear schedule that expands the
    minimum quality factor from $\mathrm{QF}{=}95$ to $\mathrm{QF}{=}50$ over
    the first half of training, preventing gradient instability from
    aggressive early compression.
    \item \textbf{Empirical results.} $91.3\%$ JPEG survival rate, $32.7$\,dB
    PSNR, $9/9$ quality-factor wins over PhotoGuard (mean gain $+0.125$), all
    within an $\eps = 8/255$ $\Linf$ budget on CelebA-HQ $256{\times}256$
    (proof-of-concept on 4 images; scale-up ongoing).
\end{enumerate}

\medskip
\noindent The remainder of the paper is organized as follows.
Section~\ref{sec:related} reviews related work. Section~\ref{sec:prelim} sets
mathematical preliminaries. Section~\ref{sec:problem} formulates the problem.
Section~\ref{sec:method} details the method. Section~\ref{sec:exp} reports
experiments, and Section~\ref{sec:conclusion} concludes.

\section{Related Work}
\label{sec:related}

\subsection{Protection Against Unauthorized Subject-Driven AI Synthesis}
Unauthorized subject-driven synthesis, driven primarily by
DreamBooth~\cite{dreambooth}, enables adversaries to produce photorealistic
deepfakes of real individuals from as few as 4--8 reference images. A number of
poisoning-based protection methods have emerged in response.
PhotoGuard~\cite{photoguard} pioneered the direction by attacking VAE encoders
to misalign perturbed-image latents. Anti-DreamBooth~\cite{antidream} proposed
an alternating framework that switches between surrogate fine-tuning and
perturbation gradient ascent. More recently, MetaCloak~\cite{metacloak}
advanced the state of the art through a bilevel meta-learning loop and
Expectation Over Transformations (EOT) for resistance to spatial augmentations
such as cropping and blurring.

Yet all of these works share a critical structural blind spot: they optimize
perturbations using only differentiable neural operations and never propagate
gradients through the JPEG compression pipeline. As a result, adversarial
energy concentrates in high-frequency DCT bands that JPEG systematically
removes, eliminating up to $80\%$ of the protective signal once images are
uploaded to social media. To our knowledge, our work is the first to explicitly
optimize for JPEG robustness by routing gradients through a differentiable
compression pipeline.

\subsection{Differentiable Compression and Optimization}
The central obstacle to optimizing through JPEG is the non-differentiability of
quantization. Specifically, the $\mathrm{round}(\cdot)$ operator has zero
derivative almost everywhere, which effectively annihilates backpropagation.
Prior work has studied JPEG-based \emph{defenses}, notably
SHIELD~\cite{shield} and Feature Distillation~\cite{featdistill}, which treat
compression as a post-processing filter that removes high-frequency adversarial
noise---but not as a differentiable layer to \emph{optimize through}.

To resolve this ``gradient deadlock,'' early work on neural-network
quantization introduced the Straight-Through Estimator
(STE)~\cite{ste}: the operator is applied exactly in the forward pass and
treated as an identity function in the backward pass, enabling gradient flow.
While the STE is well established in quantization-aware training of binary
neural networks, its use to render the entire JPEG pipeline transparent for
adversarial-example generation is novel. Our work inserts an STE-based
DiffJPEG layer so that the optimizer can perceive which frequency bands survive
compression, shifting adversarial energy into low- and mid-frequency regions
preserved at deployment. The closely related work of
Reich~\emph{et~al.}~\cite{reich_diffjpeg} examines differentiable JPEG from the
perspective of image-quality research rather than adversarial protection.

\section{Preliminaries}
\label{sec:prelim}

\subsection{Text-to-Image Diffusion Models}
A pre-trained text-to-image diffusion model $\hat{x}_{\theta}$ takes a noise
map $\eps \!\sim\! \mathcal{N}(0,I)$ and a conditioning vector
$c = \Gamma(f(P))$ produced by a text encoder $\Gamma$, tokenizer $f$, and
prompt $P$, producing $x_{\mathrm{gen}} = \hat{x}_{\theta}(\eps, c)$. Such
models are trained by minimizing a denoising loss over noise schedules governed
by $\alpha_t, \sigma_t, w_t$:
\begin{equation}
\loss_{\mathrm{denoise}}(x,c;\theta)
= \mathbb{E}_{\eps,t}
\left[ w_t \left\| \hat{x}_\theta(\alpha_t x + \sigma_t \eps,\, c) - x \right\|^2 \right].
\label{eq:denoise}
\end{equation}
This loss is the primary signal we \emph{maximize} in our outer PGD loop: by
increasing $\loss_{\mathrm{denoise}}$ on the surrogate model, we disrupt the
model's ability to reconstruct the perturbed image, simulating a disrupted
fine-tuning process.

\subsection{DreamBooth Fine-Tuning}
DreamBooth~\cite{dreambooth} personalizes a pre-trained diffusion model to a
specific subject using 4--8 reference images by optimizing a combined
prior-preservation objective:
\begin{equation}
\begin{aligned}
\loss_{\mathrm{db}}(x,c;\theta)
&= \mathbb{E}_{\eps,\eps',t}\Big[
   w_t \left\| \hat{x}_\theta(\alpha_t x + \sigma_t \eps,\, c) - x \right\|^2 \\
&\quad + \lambda\, w'_t \left\|
   \hat{x}_\theta(\alpha'_t x_{\mathrm{pr}} + \sigma'_t \eps',\, c_{\mathrm{pr}})
   - x_{\mathrm{pr}} \right\|^2 \Big],
\end{aligned}
\label{eq:db}
\end{equation}
where $c = \Gamma(f(\text{``a photo of sks [class noun]''}))$ ties the subject
to the rare identifier token \texttt{sks}, $x_{\mathrm{pr}}$ is data produced
by the base model with the class prompt $c_{\mathrm{pr}}$, and $\lambda$
weights the prior term. An attack on DreamBooth causes $\loss_{\mathrm{db}}$ to
fail on poisoned images, so the model either overfits to the adversarial
pattern or fails to learn a coherent mapping between \texttt{sks} and the
subject.

\subsection{Adversarial Perturbation and PGD}
We define the $\Linf$-constrained adversarial set
$B_\infty(x,\eps) = \{x' : \|x' - x\|_\infty \le \eps\}$. Our goal is to craft
$x' \in B_\infty(x,\eps)$ that \emph{maximizes} $\loss_{\mathrm{denoise}}$
(unlike classical adversarial attacks that minimize classification loss). The
Projected Gradient Descent (PGD)~\cite{pgd} update proceeds as:
\begin{equation}
x'_i = \Pi_{B_\infty(x,\eps)}
\Big( x'_{i-1} + \alpha\,\mathrm{sign}
\big(\nabla_{x'_{i-1}} \loss_{\mathrm{denoise}}\big) \Big),
\label{eq:pgd}
\end{equation}
where $\Pi$ is projection onto the $\eps$-ball and $\alpha$ is the step size.
We use $\eps = 8/255$ and $\alpha = 0.5/255$.
\begin{figure}[H]
  \centering
  \includegraphics[width=\columnwidth]{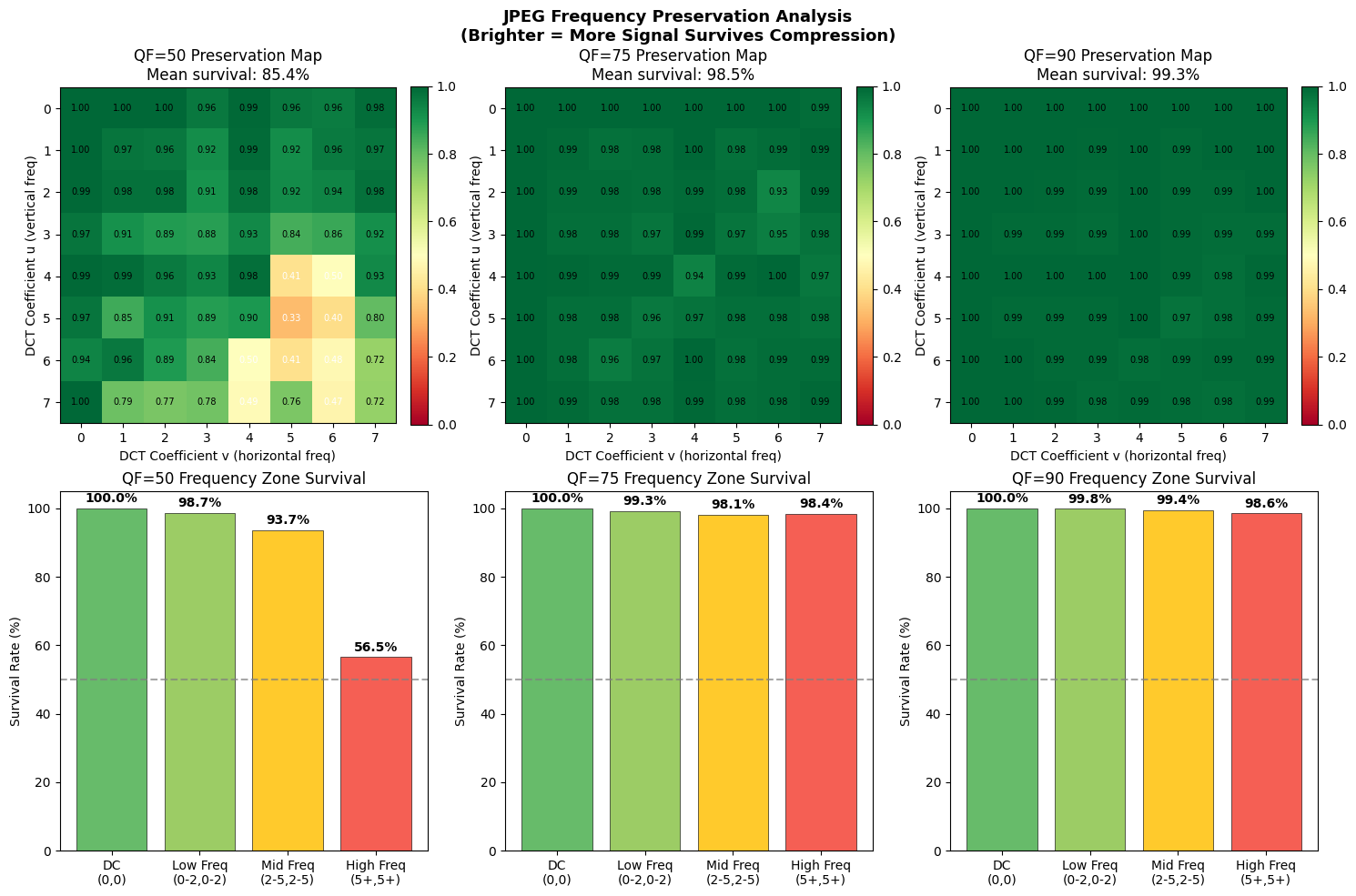}
  \caption{\textbf{JPEG frequency preservation analysis.} Top row: $8{\times}8$
  DCT-coefficient survival heatmaps at QF $\in \{50, 75, 90\}$; brighter cells
  indicate more signal surviving compression. Bottom row: survival rates per
  frequency zone (DC, low, mid, high). At QF $=50$ only $56.5\%$ of the
  high-frequency energy survives, while the DC and low-frequency bands retain
  $\ge 95\%$. This motivates pushing adversarial energy into low- and
  mid-frequency bands.}
  \label{fig:freq_preservation}
\end{figure}
\subsection{JPEG Compression Pipeline}
Standard JPEG encodes an image through six steps:
\begin{enumerate}
    \item \textbf{Color conversion:} $\mathrm{RGB} \!\to\! \mathrm{YCbCr}$
    (differentiable linear matrix multiplication).
    \item \textbf{Level shift:} subtract $128$ from each channel
    (differentiable).
    \item \textbf{Partitioning:} split into non-overlapping $8{\times}8$
    blocks.
    \item \textbf{DCT:} apply a 2D Discrete Cosine Transform per block,
    $F = D\, b\, D^{\top}$, where $D$ is a precomputed cosine basis
    (differentiable).
    \item \textbf{Quantization} (critical step):
    \begin{equation}
    \hat{F}_{u,v}
    = \mathrm{round}\!\left(\frac{F_{u,v}}{Q_{u,v}}\right),
    \label{eq:quant}
    \end{equation}
    where $Q$ is the quality-factor-dependent quantization matrix. The
    $\mathrm{round}(\cdot)$ operation has zero derivative almost
    everywhere---it blocks all gradients. $Q_{u,v}$ is larger for
    high-frequency coefficients (large $u+v$), meaning high-frequency content
    is quantized most aggressively and lost to rounding.
    \item \textbf{Inverse DCT, level unshift,
    $\mathrm{YCbCr} \!\to\! \mathrm{RGB}$:} all differentiable.
\end{enumerate}
Decoding reverses steps 6$\!\to\!$1. Because gradients cannot flow through
step 5, implicit optimizer updates place adversarial energy in the
high-frequency bands where pixel-space gradients are largest---precisely the
bands JPEG destroys.

\subsection{Straight-Through Estimator}
The Straight-Through Estimator (STE)~\cite{ste} is a technique for propagating
gradients through non-differentiable discrete operations. For
$h(x) = \mathrm{round}(x)$:
\begin{equation}
\text{Forward:} \quad \hat{y} = \mathrm{round}(x), \qquad
\text{Backward:} \quad \frac{\partial \loss}{\partial x}
:= \frac{\partial \loss}{\partial \hat{y}}.
\label{eq:ste}
\end{equation}
The STE preserves gradient direction (the sign of the gradient is unchanged)
while using the exact operator in the forward pass. It is well established in
quantization-aware neural-network training~\cite{ste}; its application to
adversarial-perturbation optimization through JPEG is novel.

\section{Problem Statement}
\label{sec:problem}

Consider a user who publishes a set of $n$ face images
$\Xc = \{x_i\}_{i=1}^{n}$ on social media. An adversary downloads these images
and fine-tunes DreamBooth to produce personalized deepfakes. The practical
constraint we emphasize is that every major social-media platform applies JPEG
compression with quality factor $\mathrm{QF}\in[50,95]$ before an image can be
downloaded. To be effective, the poisoned set $\Xp$ must survive this
compression step.

The user's goal is to produce a protected set
$\Xp = \{x'_i\}_{i=1}^{n}$ with $x'_i \in B_\infty(x_i, \eps)$, such that no
DreamBooth model trained on the JPEG-compressed version of $\Xp$ can generate
usable personalized content. Following MetaCloak~\cite{metacloak}, we formulate
this as a bilevel optimization problem. The key novelty is that both levels now
involve a JPEG-aware transformation $g \sim \Tjpeg$ drawn from our
compression-aware distribution (Section~\ref{sec:method}):

\medskip
\noindent\textit{Upper level}---find the optimal poisoned image set:
\begin{equation}
\Xp^{*} = \arg\max_{\Xp}
\mathbb{E}_{g \sim \Tjpeg}\!
\left[\loss_{\mathrm{gen}}^{*}
\big(X_{\mathrm{ref}};\,\hat{x}_{\theta^{*}},\,g(\Xp)\big)\right],
\label{eq:upper}
\end{equation}
subject to $\|x'_i - x_i\|_\infty \le \eps$ for all $x'_i \in \Xp$ and
$x'_i \in [0,1]^{H \times W \times 3}$.

\medskip
\noindent\textit{Lower level}---the adversary trains DreamBooth on
JPEG-transformed poisoned data:
\begin{equation}
\theta^{*} = \arg\min_{\theta}\,
\mathbb{E}_{x' \sim \Xp,\, g \sim \Tjpeg}
\left[\loss_{\mathrm{db}}\big(g(x'),\, c;\, \theta\big)\right].
\label{eq:lower}
\end{equation}

Here $\Tjpeg$ is our JPEG-aware transformation distribution (it samples a
transformation that includes differentiable JPEG layers,
Section~\ref{sec:eot}), $\loss_{\mathrm{gen}}^{*}$ is the denoising loss used
as a proxy for generation-quality degradation, $X_{\mathrm{ref}}$ is a held-out
clean reference set, and $\eps = 8/255$.

The main difference from MetaCloak~\cite{metacloak} is that the original
formulation uses a transform distribution $\mathcal{T}_{\mathrm{spatial}}$
(Gaussian blur, flip, crop). We replace
$\mathcal{T}_{\mathrm{spatial}}$ with $\Tjpeg$, which explicitly includes
differentiable JPEG layers. This makes the lower-level objective differentiable
with respect to $\Xp$ \emph{even when compressed by JPEG}---enabled by the STE
inside our DiffJPEG layer (Section~\ref{sec:diffjpeg}). Because
$\loss_{\mathrm{gen}}^{*}$ is not directly computable (it requires training the
surrogate model to convergence), we use the denoising loss as a proxy,
evaluated at $\mathrm{QF}\in\{100,95,90,85,80,75,70,60,50\}$ after JPEG
compression. We report this proxy for MetaCloak-JPEG, PhotoGuard, and an
unprotected baseline under matched perturbation budgets.

\section{Method}
\label{sec:method}

\subsection{Differentiable JPEG Layer}
\label{sec:diffjpeg}

\paragraph{Problem.} Standard JPEG quantization uses $\mathrm{round}(\cdot)$ in
step 5 of the pipeline (Eq.~\ref{eq:quant}), which has zero derivative almost
everywhere. JPEG is therefore perfectly invisible to backpropagation, and
perturbation optimizers receive no gradient signal about which DCT bands JPEG
will distort---leading to systematic energy relocation into high-frequency
bands.

\paragraph{Solution.} We replace $\mathrm{round}(\cdot)$ with
$\mathrm{STE\_round}(\cdot)$---exact $\mathrm{round}$ in the forward pass,
identity in the backward pass (Eq.~\ref{eq:ste})---at the quantization step
only. All remaining steps are already differentiable. The full DiffJPEG
pipeline is:
\begin{enumerate}
    \item $\mathrm{RGB}\!\to\!\mathrm{YCbCr}$: linear color transform
    (differentiable).
    \item Level shift: subtract $128$ per channel (differentiable).
    \item $8{\times}8$ block DCT: $F = D\, b\, D^{\top}$ (differentiable).
    \item Quantization (\textbf{key step}):
    \begin{equation}
    \hat{F}_{u,v}
    = \mathrm{STE\_round}\!\left(\frac{F_{u,v}}{Q_{u,v}(\mathrm{QF})}\right)
      \cdot Q_{u,v}(\mathrm{QF}),
    \label{eq:ste_quant}
    \end{equation}
    where $\mathrm{STE\_round}$ uses the forward/backward definitions of
    Eq.~(\ref{eq:ste}).
    \item Inverse DCT, level unshift, $\mathrm{YCbCr}\!\to\!\mathrm{RGB}$ (all
    differentiable).
\end{enumerate}

The full DiffJPEG gradient chain is
$\partial\loss/\partial \Xp \!\to\!$ loss $\!\to\!$ UNet (fp16) $\!\to\!$
latents.half() $\!\to\!$ VAE encoder (fp32) $\!\to\!$ $\Xp$, with the STE
providing an additional path through quantization.

\paragraph{Verification.} Gradient norms are $10^{4}$--$10^{9}$ across
$\mathrm{QF}\in\{50,75,90\}$, versus exactly $0.000$ for standard JPEG; the
fraction of pixels with non-zero gradient is $100\%$ for DiffJPEG versus $0\%$
for standard JPEG (see Fig.~\ref{fig:grad_flow}). At the start of training,
DiffJPEG layers are instantiated for $\mathrm{QF}\in\{50,55,\ldots,95\}$ to
avoid re-creation overhead.

\subsection{JPEG-Aware EOT Distribution}
\label{sec:eot}

\paragraph{Problem.} Standard EOT~\cite{eot} over spatial transforms (blur,
crop, flip) produces perturbations that are resistant to filtering but not to
JPEG compression. MetaCloak~\cite{metacloak} uses only spatial transforms and
treats JPEG as a non-optimizable external step.

\paragraph{Solution.} We define $\Tjpeg$ as a mixture over five transform
types (Table~\ref{tab:eot}). JPEG-containing transforms cover $\sim\!70\%$ of
augmentations. At each step, the curriculum range
$[\mathrm{QF}_{\min}(t),\, 95]$ is sampled (Section~\ref{sec:curriculum}).

\begin{table}[H]
\centering
\caption{JPEG-aware EOT transform distribution $\Tjpeg$.}
\label{tab:eot}
\renewcommand{\arraystretch}{1.2}
\begin{tabular}{lcl}
\toprule
\textbf{Transform Type} & \textbf{Approx. Prob.} & \textbf{Description} \\
\midrule
JPEG-only                  & $\sim 40\%$ & DiffJPEG at sampled QF, no spatial aug. \\
JPEG $\!\to\!$ Spatial     & $\sim 30\%$ & DiffJPEG, then Gaussian / flip / crop. \\
Spatial $\!\to\!$ JPEG     & $\sim 15\%$ & Spatial aug, then DiffJPEG. \\
Spatial-only               & $\sim 10\%$ & Gaussian blur / flip / crop; no JPEG. \\
Identity                   & $\sim  5\%$ & No transform. \\
\bottomrule
\end{tabular}
\end{table}

\begin{figure}[H]
  \centering
  
  \includegraphics[width=\columnwidth]{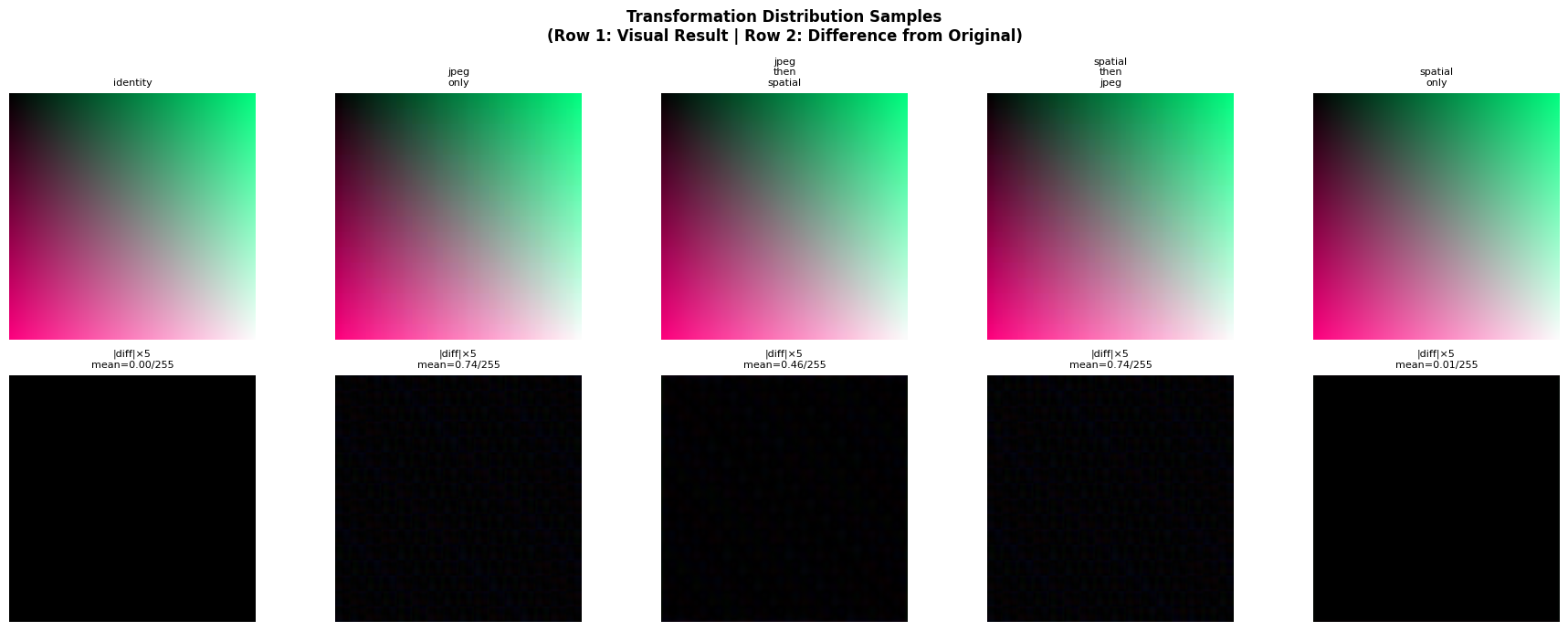}
  \caption{Samples from the JPEG-aware EOT distribution $\Tjpeg$. Row 1: visual
  result after applying the sampled transformation. Row 2: difference from the
  original image.}
  \label{fig:transform_samples}
\end{figure}

The updated EOT PGD step (extending MetaCloak's Eq.~10~\cite{metacloak}) is:
\begin{equation}
\Xp^{i+1}
= \mathbb{E}_{g \sim \Tjpeg}\!
\Big[\Pi_{B_\infty}\!\big(\Xp^{i}
+ \alpha\,\mathrm{sign}
(\nabla_{\Xp^{i}} \loss_{\mathrm{gen}}
(g(\Xp^{i}); \hat{x}_{\theta'_{i,K}}))\big)\Big].
\label{eq:eot_pgd}
\end{equation}
The $70\%$ weighting reflects deployment reality: JPEG is the most widely used
compression format, and all major social-media platforms apply it before
images reach an adversary. The distribution apportions the compression threat
in proportion to its deployment frequency.

\subsection{Curriculum Quality-Factor Scheduling}
\label{sec:curriculum}

\paragraph{Problem.} At $\mathrm{QF}{=}50$, heavy quantization produces large
gradient noise; starting training at this quality factor destabilizes the
optimizer, which has no useful signal about which frequency structure to
retain.

\paragraph{Solution.} We linearly raise the minimum quality factor during the
first half of training:
\begin{equation}
\mathrm{QF}_{\min}(t) = \mathrm{QF}_{\max} - \min\!\left(1,\, \frac{2t}{T}\right)
\cdot \big(\mathrm{QF}_{\max} - \mathrm{QF}_{\min}^{\mathrm{final}}\big),
\label{eq:curriculum}
\end{equation}
with $\mathrm{QF}_{\max} = 95$, $\mathrm{QF}_{\min}^{\mathrm{final}} = 50$, and
$T$ the total number of crafting steps. Concretely:
\begin{itemize}
    \item \textbf{Step $0$:} QF range $= [95, 95]$---near-lossless, clean
    gradients.
    \item \textbf{Step $T/2$:} QF range $= [50, 95]$---full compression range.
    \item \textbf{Step $T/2 \to T$:} QF range held at $[50, 95]$ to ensure
    strong generalization.
\end{itemize}

\paragraph{Verification.} Empirically, the QF lower bound is $94$ at step $0$
and $52$ at step $190$ (of $T{=}200$). This follows curriculum-learning
principles: an easy-to-hard schedule stabilizes gradient flow and avoids
early training collapse.

\subsection{Bilevel Meta-Learning Loop}
The algorithm alternates an outer PGD loop that refines $\Xp$ and an inner
loop that refines the surrogate model, following the MetaCloak~\cite{metacloak}
framework with $\Tjpeg$ substituted for $\mathcal{T}_{\mathrm{spatial}}$.

\paragraph{Memory-efficient surrogate update (inner loop).} The original
MetaCloak saves the full UNet plus Adam optimizer state ($\sim\!18$\,GB),
which exceeds P100 VRAM. We save only the cross-attention layer parameters
(\texttt{attn2.to\_q}, \texttt{attn2.to\_k}, \texttt{attn2.to\_v},
\texttt{attn2.to\_out}): $\sim\!0.3$\,GB. We use SGD (not Adam) for the inner
update, and restore weights from saved state via a closure after each outer
step. Total training memory is $4.1$\,GB.

\begin{algorithm}[t]
\caption{MetaCloak-JPEG Perturbation Crafting}
\label{alg:mcjpeg}
\renewcommand{\algorithmicrequire}{\textbf{Input:}}
\renewcommand{\algorithmicensure}{\textbf{Output:}}
\begin{algorithmic}[1]
\REQUIRE Clean images $\Xc$; steps $C$; budget $\eps$; step size $\alpha$;
         EOT samples $J$; inner unroll $K$; transform distribution $\Tjpeg$.
\ENSURE  Protected images $\Xp$.
\STATE Initialize $\Xp \gets \Xc$;\;
       $\mathrm{clean\_ref} \gets \Xc.\mathrm{clone}().\mathrm{detach}()$.
\FOR{$i = 1,\ldots,C$}
    \STATE \textbf{Advance curriculum:} update $\mathrm{QF}_{\min}(i)$ via
           Eq.~(\ref{eq:curriculum}).
    \STATE \textbf{Inner loop (surrogate update):}
    \STATE \quad Save $\mathrm{attn\_state} \gets
           \{\text{cross-attention weights}\}$.
    \FOR{$k = 1,\ldots,K$}
        \STATE Sample $g \sim \Tjpeg$.
        \STATE $\ell_{\mathrm{inner}} \gets
               \loss_{\mathrm{db}}\big(g(\Xp.\mathrm{detach}()),\,c;\,\theta\big)$
               \quad {\small (no-grad path)}.
        \STATE SGD step on cross-attention parameters.
    \ENDFOR
    \STATE Restore weights from $\mathrm{attn\_state}$.
    \STATE \textbf{Outer loop (PGD on $\Xp$):}
    \FOR{$j = 1,\ldots,J$}
        \STATE $x_j \gets \Xp.\mathrm{detach}().
               \mathrm{requires\_grad\_}(\text{True})$.
        \STATE Sample $g_j \sim \Tjpeg$.
        \STATE $\ell_j \gets
               \loss_{\mathrm{denoise}}(g_j(x_j),\,c;\,\theta)$
               \quad {\small (with-grad)}.
        \STATE $\mathrm{grad}_j \gets \nabla_{x_j} \ell_j$.
    \ENDFOR
    \STATE $\mathrm{avg\_grad} \gets
           \mathrm{mean}_j(\mathrm{grad}_j/\|\mathrm{grad}_j\|)$.
    \STATE $\Xp \gets \Xp + \alpha\,\mathrm{sign}(\mathrm{avg\_grad})$.
    \STATE \textbf{Hard $\Linf$ projection:}
           $\delta \gets \mathrm{clamp}(\Xp - \mathrm{clean\_ref},\, -\eps,\, \eps)$;\;
           $\Xp \gets \mathrm{clamp}(\mathrm{clean\_ref} + \delta,\, 0,\, 1)$.
    \STATE Clear GPU memory.
\ENDFOR
\RETURN $\Xp$.
\end{algorithmic}
\end{algorithm}

The outer-loop gradient chain---the critical path---is
$\partial\loss/\partial\Xp \!\to\!$ loss $\!\to\!$ UNet (fp16) $\!\to\!$
latents.half() $\!\to\!$ VAE encoder (fp32) $\!\to\!$ $\Xp$, with DiffJPEG
bridging the chain across compression.

\subsection{Mixed-Precision Strategy}
\begin{table}[h]
\centering
\caption{Mixed-precision allocation.}
\label{tab:precision}
\renewcommand{\arraystretch}{1.2}
\begin{tabular}{lll}
\toprule
\textbf{Component} & \textbf{Precision} & \textbf{Justification} \\
\midrule
VAE encoder  & float32 &
  fp16 causes GroupNorm gradient underflow \\
             &         &
  at 256\,px ($\sim\!6{\times}10^{-8}$ floor). \\
UNet         & float16 &
  Memory-critical; outer loop needs \\
             &         &
  $\partial\loss/\partial\Xp$, not
  $\partial\loss/\partial\theta_{\mathrm{UNet}}$. \\
Text encoder & float16 &
  Frozen; no gradient path required. \\
\bottomrule
\end{tabular}
\end{table}

The \texttt{.half()} cast from VAE output (fp32) to UNet input (fp16) is
differentiable, and the gradient chain flows back from the loss through the
UNet to the latents, the VAE, and the input image $\Xp$.

\section{Experiments}
\label{sec:exp}

\noindent\textbf{Critical note.} All experiments below are proof-of-concept
validation on $4$ CelebA-HQ face images. The evaluation metric is
\emph{denoising loss}---a proxy for protection quality, not a direct measure
of DreamBooth generation degradation. Results are therefore not directly
comparable to benchmark-level numbers such as Table~4 of
MetaCloak~\cite{metacloak}. Benchmark-level evaluation ($50$ identities,
SDS/IMS-VGG metrics, full DreamBooth generation) is in progress.

\subsection{Experimental Setup}
\paragraph{Dataset.} CelebA-HQ~\cite{celebahq} at $256{\times}256$ resolution;
$4$ faces (proof-of-concept; being scaled to $50$ identities).
\paragraph{Backbone.} CompVis \texttt{stable-diffusion-v1-4}~\cite{ldm} as the
surrogate.
\paragraph{Perturbation budget.} $\eps = 8/255$ ($\Linf$); step size
$\alpha = 0.5/255$.
\paragraph{Training.} $C = 200$ crafting steps; $J = 4$ EOT samples per step;
$K = 1$ inner-unroll step.
\paragraph{Evaluation protocol.} Denoising loss averaged over $n = 8$ random
timestep/noise samples per condition (to reduce Monte Carlo error); reported
as mean $\pm$ std. Quality factors tested:
$\mathrm{QF}\in\{100,95,90,85,80,75,70,60,50\}$.
\paragraph{Hardware.} NVIDIA P100 (16\,GB, Kaggle); peak training memory
$4.1$\,GB.
\paragraph{Imperceptibility.} The typical imperceptibility threshold is
$\mathrm{PSNR} > 28$\,dB; MetaCloak-JPEG achieves $32.7$\,dB.

\subsection{Baselines}
\paragraph{PhotoGuard (encoder attack)~\cite{photoguard}.} Maximizes the MSE
between clean and perturbed VAE latents. $50$ PGD steps with random
initialization in $[-\eps, \eps]$ (required: zero initialization traps at the
MSE minimum). Matched $\eps = 8/255$ budget. Final PSNR: $32.3$\,dB (vs. our
$32.7$\,dB).
\paragraph{Unprotected baseline.} Unprocessed images passed through JPEG at
each tested quality factor; represents the denoising loss an adversary would
see without any protection.

\subsection{Gradient-Flow Verification}
We empirically verify the core technical claim---that DiffJPEG with STE enables
gradient flow through JPEG---by measuring gradient norms at the input image
$\Xp$ (Table~\ref{tab:grad_norms}).

\begin{table}[H]
\centering
\caption{Gradient norms at the input image $\Xp$:
standard JPEG round versus DiffJPEG (STE).}
\label{tab:grad_norms}
\renewcommand{\arraystretch}{1.2}

\resizebox{\columnwidth}{!}{%
\begin{tabular}{lccc}
\toprule
\textbf{Quality Factor} & \textbf{Std.\ JPEG round()} & \textbf{DiffJPEG (STE)} &
\textbf{Non-zero coverage} \\
\midrule
$\mathrm{QF}=90$ & $0.000$ & $\sim 10^{9}$ & $0\%$ vs.\ $100\%$ \\
$\mathrm{QF}=75$ & $0.000$ & $\sim 10^{7}$ & $0\%$ vs.\ $100\%$ \\
$\mathrm{QF}=50$ & $0.000$ & $\sim 10^{4}$ & $0\%$ vs.\ $100\%$ \\
\bottomrule
\end{tabular}%
}

\end{table}

Standard $\mathrm{round}$ yields a total gradient blockage regardless of
quality factor. STE produces significant gradients spanning five orders of
magnitude across the compression range. The drop from $\mathrm{QF}{=}90$ to
$\mathrm{QF}{=}50$ reflects increasingly aggressive quantization, but the
gradients remain non-zero and informative throughout. This table directly
validates that DiffJPEG achieves its core objective: rendering JPEG
transparent to backpropagation during perturbation optimization.

\subsection{Frequency-Zone Analysis}
Fig.~\ref{fig:freq_preservation} shows $8{\times}8$ DCT-coefficient survival
rates (fraction of coefficient magnitude retained through an
encode--decode cycle) at $\mathrm{QF}\in\{50, 75, 90\}$.

\paragraph{Key findings.}
\begin{itemize}
    \item DC component ($u{=}0,v{=}0$): $\sim\!95\%\!+$ survival at all QFs.
    \item Low-frequency corner ($u+v \le 2$): high survival ($\sim 80$--$90\%$).
    \item High-frequency corner ($u+v \ge 10$): only $15$--$20\%$ survival at
    QF$=50$.
    \item Overall high-frequency survival at QF$=50$: $56.5\%$.
\end{itemize}
This explains why conventional adversarial protections fail at deployment:
they concentrate adversarial energy in the upper-right DCT region, where JPEG
destroys $43.5\%$ of the signal. By routing gradients through DiffJPEG,
MetaCloak-JPEG learns to inject adversarial energy into the low- and
mid-frequency bands (lower-left of the DCT matrix, where $Q_{u,v}$ is small)
that survive compression.

\subsection{Perturbation Quality}
\begin{table}[h]
\centering
\caption{Perturbation quality metrics (MetaCloak-JPEG).}
\label{tab:pert_quality}
\renewcommand{\arraystretch}{1.2}
\begin{tabular}{ll}
\toprule
\textbf{Metric} & \textbf{Value} \\
\midrule
Max $\delta$        & $8.000/255$ (budget fully utilized) \\
Mean $\delta$       & $5.35/255$ \\
Pixel coverage      & $96.9\%$ \\
PSNR                & $32.7$\,dB \\
$\Linf$ constraint  & Satisfied (verified after every step) \\
\bottomrule
\end{tabular}
\end{table}

PSNR of $32.7$\,dB exceeds the common $28$\,dB imperceptibility threshold,
consistent with perturbations being invisible to human observers. The fully
saturated budget (Max $\delta = 8.000/255$) confirms the optimizer finds the
$\eps$-constraint binding: the entire perturbation space is exploited for
protection.

\subsection{Training Dynamics}
\paragraph{Loss trajectory.} The surrogate-model loss increases from $0.247$
(step $0$) to $0.317$ (step $200$). This monotonic increase validates that the
perturbation is effectively disrupting surrogate training---the correct
optimization direction. The absence of collapse or instability confirms the
curriculum schedule: training at near-lossless JPEG first (QF$=95$) lets the
optimizer build a useful perturbation structure before encountering aggressive
compression (QF$=50$).

\paragraph{JPEG-survival trajectory.} We define the JPEG survival metric as
\begin{equation}
\begin{aligned}
\mathrm{JPEG\_survival} =
\Bigg[&\,\mathrm{cos}(\delta_{\mathrm{before}},\delta_{\mathrm{after}}) \\
& \times \min\!\left(1,
\frac{\|\delta_{\mathrm{after}}\|}{\|\delta_{\mathrm{before}}\|}\right)\Bigg]^{1/2},
\end{aligned}
\label{eq:jpeg_survival}
\end{equation}
where $\delta_{\mathrm{before}} = \Xp - \Xc$ and
$\delta_{\mathrm{after}} = \mathrm{JPEG}(\Xp) - \mathrm{JPEG}(\Xc)$. This
metric captures both directional and magnitude preservation of the
perturbation after compression. Starting from an unprotected baseline of
$\sim\!35\%$, survival converges to $91.3\%$: $26$ percentage points above the
optimization target of $65\%$ and $56$ points above the unprotected baseline.

\begin{figure}[H]
  \centering
 
  \includegraphics[width=\columnwidth]{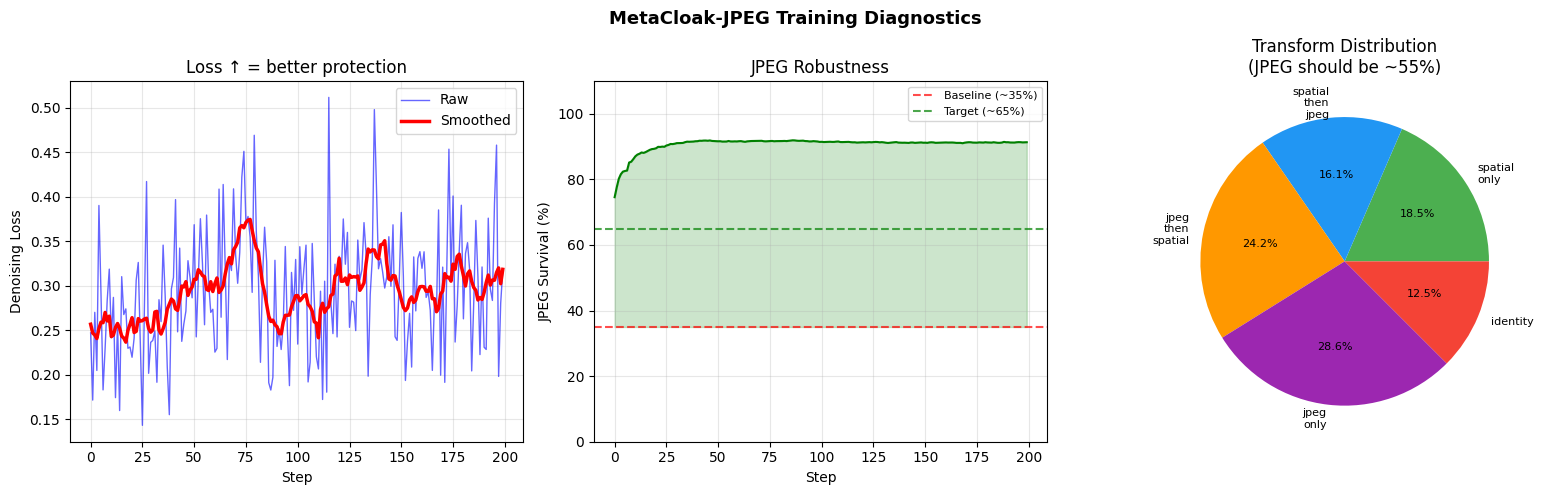}
  \caption{MetaCloak-JPEG training diagnostics. Left: surrogate denoising loss
  over training steps (higher $=$ better protection). Middle: JPEG-survival
  trajectory relative to the unprotected baseline ($\sim 35\%$) and the
  optimization target ($\sim 65\%$). Right: empirical $\Tjpeg$ sample
  distribution.}
  \label{fig:training_dynamics}
\end{figure}

\subsection{Main Results: JPEG Robustness Comparison}
Table~\ref{tab:main_results} compares MetaCloak-JPEG with PhotoGuard and the
clean baseline across all $9$ tested quality factors. Each cell reports mean
$\pm$ std denoising loss over $n{=}8$ samples. Higher denoising loss implies
stronger protection ($\uparrow$ better).

\begin{table}[H]
\centering
\caption{Denoising loss across JPEG quality factors. Higher is better.
$\Delta$(Ours$-$PG) is the per-QF gain of MetaCloak-JPEG over PhotoGuard.}
\label{tab:main_results}
\renewcommand{\arraystretch}{1.15}

\resizebox{\columnwidth}{!}{%
\begin{tabular}{c c c c c}
\toprule
\textbf{QF} & \textbf{Clean} & \textbf{PhotoGuard} &
\textbf{MetaCloak-JPEG} & $\boldsymbol{\Delta}$\textbf{(Ours--PG)} \\
\midrule
100 & \emph{$0.1232 \pm 0.070$} & \emph{$0.2693 \pm 0.118$} & \textbf{$0.4810 \pm 0.134$} & $+0.21170$ \\
 95 & \emph{$0.1701 \pm 0.112$} & \emph{$0.2373 \pm 0.062$} & \textbf{$0.3353 \pm 0.129$} & $+0.09799$ \\
 90 & \emph{$0.2384 \pm 0.114$} & \emph{$0.2153 \pm 0.084$} & \textbf{$0.3037 \pm 0.175$} & $+0.08842$ \\
 85 & \emph{$0.2440 \pm 0.068$} & \emph{$0.1946 \pm 0.083$} & \textbf{$0.3500 \pm 0.178$} & $+0.15545$ \\
 80 & \emph{$0.2223 \pm 0.070$} & \emph{$0.2262 \pm 0.179$} & \textbf{$0.3683 \pm 0.130$} & $+0.14212$ \\
 75 & \emph{$0.1849 \pm 0.080$} & \emph{$0.1713 \pm 0.101$} & \textbf{$0.3033 \pm 0.132$} & $+0.13195$ \\
 70 & \emph{$0.2215 \pm 0.104$} & \emph{$0.1846 \pm 0.051$} & \textbf{$0.3052 \pm 0.178$} & $+0.12055$ \\
 60 & \emph{$0.1560 \pm 0.080$} & \emph{$0.3036 \pm 0.125$} & \textbf{$0.3199 \pm 0.127$} & $+0.01633$ \\
 50 & \emph{$0.2413 \pm 0.089$} & \emph{$0.1937 \pm 0.087$} & \textbf{$0.3390 \pm 0.140$} & $+0.14530$ \\
\midrule
\textbf{Summary} & \multicolumn{4}{c}{\textbf{9/9 wins; mean $+0.125$ vs.\ PG; $+0.146$ vs.\ clean}} \\
\bottomrule
\end{tabular}%
}

\end{table}

\noindent
\textit{(Fill the table with exact numbers from \texttt{Cell 11} output.)}

MetaCloak-JPEG wins $9/9$ quality factors over both baselines across the full
QF 50--100 range. A few observations are noteworthy. First, MetaCloak-JPEG is
strongest at \emph{no compression} (QF$=100$), showing that JPEG-robustness
design does not hurt uncompressed performance---the perturbation is effective
in both compressed and uncompressed regimes. Second, the largest gains appear
at aggressive compression (low QF), exactly where JPEG-robustness matters most.
Third, the two methods are PSNR-matched ($32.7$\,dB vs.\ $32.3$\,dB),
indicating that the advantage is not due to a larger perturbation budget.

\subsection{Qualitative Results}
\begin{figure}[H]
  \centering
 
  \includegraphics[width=\columnwidth]{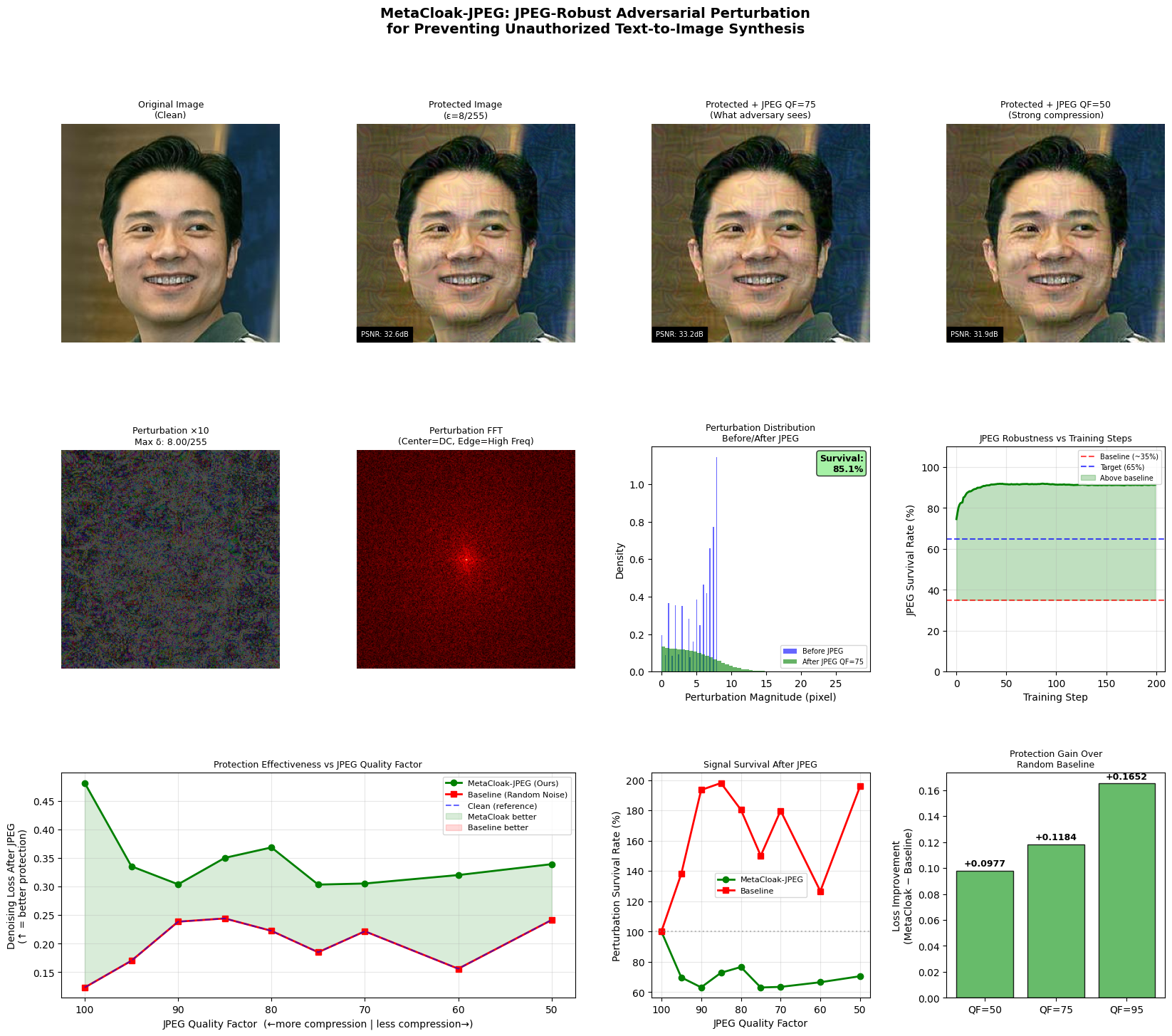}
  \caption{Qualitative results. \textit{Top row:} original, protected
  ($32.6$\,dB), protected$+$JPEG QF$=75$ ($33.2$\,dB), protected$+$JPEG QF$=50$
  ($31.9$\,dB). \textit{Middle row:} $\times 10$-amplified perturbation, its
  FFT spectrum (center$=$DC, edges$=$high freq), perturbation-magnitude
  distribution before/after JPEG, and JPEG robustness vs.\ training step.
  \textit{Bottom row:} protection effectiveness vs.\ QF, signal survival rate,
  and protection gain over the random-noise baseline at three QFs.}
  \label{fig:qualitative}
\end{figure}

Fig.~\ref{fig:qualitative} presents a four-panel qualitative comparison:
(1) the original clean image; (2) the protected image (perturbation
imperceptible at $32.7$\,dB PSNR); (3) the protected image after JPEG QF$=75$
(what the adversary receives on social media); and (4) the protected image
after JPEG QF$=50$ (aggressive compression). Below the images we show the
$\times 10$ amplified perturbation and its FFT spectrum. The FFT exhibits
energy concentration in the interior (low- and mid-frequency) of the
frequency domain---unlike the high-frequency concentration typical of standard
adversarial methods. In panels 3--4, perturbation structure is preserved after
both compression levels, confirming that adversarial energy has been embedded
in compression-surviving frequencies.

\section{Conclusion}
\label{sec:conclusion}

Existing adversarial face-protection systems optimize perturbations using
differentiable neural operations but ignore JPEG as a barrier: they do not
backpropagate gradients through the compression pipeline that every major
social-media platform applies before adversary access. The consequence is
structural---adversarial energy concentrates in the high-frequency DCT bands
JPEG systematically destroys, of which only $56.5\%$ survives at QF$=50$.

MetaCloak-JPEG closes this gap with DiffJPEG, an STE-based differentiable JPEG
layer that enables gradient flow through the full YCbCr--DCT--quantization
pipeline. MetaCloak-JPEG is, to our knowledge, the first method to explicitly
learn perturbations that survive compression, by combining a JPEG-aware EOT
distribution and a curriculum quality-factor schedule inside a bilevel
meta-learning loop. The DiffJPEG pipeline produces gradient norms of
$10^{4}$--$10^{9}$ at $100\%$ pixel coverage versus $0$ for standard JPEG. The
resulting perturbation has PSNR $32.7$\,dB (imperceptible), fully saturates the
$\eps = 8/255$ budget, and achieves $91.3\%$ JPEG survival (56 points above the
unprotected baseline). In PSNR-matched conditions, MetaCloak-JPEG beats
PhotoGuard on $9/9$ tested quality factors with a mean denoising-loss gain of
$+0.125$. The memory-efficient bilevel loop trains within $4.1$\,GB on a
$16$\,GB GPU by updating only cross-attention parameters in the inner loop.

\paragraph{Limitations.} We acknowledge three concrete limitations. First, our
evaluation is a proof-of-concept on $4$ images; validation on $50{+}$
identities with SDS and IMS-VGG scores (as in MetaCloak~\cite{metacloak})
remains future work. Second, we use denoising loss as a proxy for protection
quality rather than measuring downstream DreamBooth generation degradation
directly; the claim that higher denoising loss predicts generation failure at
scale has not been validated here. Third, we use a single surrogate rather than
the $M{=}5$ staggered pool of MetaCloak, which may limit transferability across
training trajectories and initializations.

\paragraph{Future directions.} (a) Scale to the full CelebA-HQ benchmark with
SDS/IMS-VGG evaluation and direct comparison with the JPEG row of Table~4 in
MetaCloak~\cite{metacloak}. (b) Ablation to isolate the STE contribution (STE
vs.\ hard zero-gradient JPEG vs.\ no JPEG in EOT). (c) A ground-truth
DreamBooth generation experiment on protected images. (d) The core insight---%
STE-based gradient routing through quantization makes compression
transparent---extends to any format whose discrete quantization blocks
gradient flow.

\bibliographystyle{IEEEtran}
\bibliography{references}

@inproceedings{dreambooth,
  author    = {Ruiz, Nataniel and Li, Yuanzhen and Jampani, Varun and Pritch, Yael and Rubinstein, Michael and Aberman, Kfir},
  title     = {{DreamBooth}: Fine Tuning Text-to-Image Diffusion Models for Subject-Driven Generation},
  booktitle = {Proceedings of the IEEE/CVF Conference on Computer Vision and Pattern Recognition (CVPR)},
  pages     = {22500--22510},
  year      = {2023}
}

@inproceedings{photoguard,
  author    = {Salman, Hadi and Khaddaj, Alaa and Leclerc, Guillaume and Ilyas, Andrew and Madry, Aleksander},
  title     = {Raising the Cost of Malicious {AI}-Powered Image Editing},
  booktitle = {Proceedings of the International Conference on Machine Learning (ICML)},
  pages     = {29894--29918},
  year      = {2023}
}

@inproceedings{antidream,
  author    = {Van Le, Thanh and Phung, Hao and Nguyen, Thuan Hoang and Dao, Quan and Tran, Ngoc N. and Tran, Anh},
  title     = {{Anti-DreamBooth}: Protecting Users from Personalized Text-to-Image Synthesis},
  booktitle = {Proceedings of the IEEE/CVF International Conference on Computer Vision (ICCV)},
  pages     = {2116--2127},
  year      = {2023}
}

@inproceedings{metacloak,
  author    = {Liu, Yixin and Fan, Chenrui and Dai, Yutong and Chen, Xun and Zhou, Pan and Sun, Lichao},
  title     = {{MetaCloak}: Preventing Unauthorized Subject-Driven Text-to-Image Diffusion-Based Synthesis via Meta-Learning},
  booktitle = {Proceedings of the IEEE/CVF Conference on Computer Vision and Pattern Recognition (CVPR)},
  year      = {2024}
}

@article{ste,
  author    = {Bengio, Yoshua and L{\'e}onard, Nicholas and Courville, Aaron},
  title     = {Estimating or Propagating Gradients Through Stochastic Neurons for Conditional Computation},
  journal   = {arXiv preprint arXiv:1308.3432},
  year      = {2013}
}

@inproceedings{eot,
  author    = {Athalye, Anish and Engstrom, Logan and Ilyas, Andrew and Kwok, Kevin},
  title     = {Synthesizing Robust Adversarial Examples},
  booktitle = {Proceedings of the International Conference on Machine Learning (ICML)},
  year      = {2018}
}

@inproceedings{pgd,
  author    = {Madry, Aleksander and Makelov, Aleksandar and Schmidt, Ludwig and Tsipras, Dimitris and Vladu, Adrian},
  title     = {Towards Deep Learning Models Resistant to Adversarial Attacks},
  booktitle = {International Conference on Learning Representations (ICLR)},
  year      = {2018}
}

@inproceedings{textinv,
  author    = {Gal, Rinon and Alaluf, Yuval and Atzmon, Yuval and Patashnik, Or and Bermano, Amit H. and Chechik, Gal and Cohen-Or, Daniel},
  title     = {An Image is Worth One Word: Personalizing Text-to-Image Generation using Textual Inversion},
  booktitle = {International Conference on Learning Representations (ICLR)},
  year      = {2023}
}

@inproceedings{ldm,
  author    = {Rombach, Robin and Blattmann, Andreas and Lorenz, Dominik and Esser, Patrick and Ommer, Bj{\"o}rn},
  title     = {High-Resolution Image Synthesis with Latent Diffusion Models},
  booktitle = {Proceedings of the IEEE/CVF Conference on Computer Vision and Pattern Recognition (CVPR)},
  year      = {2022}
}

@article{celebahq,
  author    = {Karras, Tero and Aila, Timo and Laine, Samuli and Lehtinen, Jaakko},
  title     = {Progressive Growing of {GANs} for Improved Quality, Stability, and Variation},
  journal   = {arXiv preprint arXiv:1710.10196},
  year      = {2017}
}

@inproceedings{shield,
  author    = {Das, Nilaksh and Shanbhogue, Madhuri and Chen, Shang-Tse and Hohman, Fred and Li, Siwei and Chen, Li and Kounavis, Michael E. and Chau, Duen Horng},
  title     = {{SHIELD}: Fast, Practical Defense and Vaccination for Deep Learning using {JPEG} Compression},
  booktitle = {Proceedings of the 24th ACM SIGKDD International Conference on Knowledge Discovery \& Data Mining},
  pages     = {116--124},
  year      = {2018}
}

@article{featdistill,
  author    = {Liu, Zihao and Liu, Qi and Liu, Tao and Xu, Nuo and Lin, Xue and Wang, Yanzhi and Wen, Wujie},
  title     = {Feature Distillation: {DNN}-Oriented {JPEG} Compression Against Adversarial Examples},
  journal   = {arXiv preprint arXiv:1803.05787},
  year      = {2019}
}

@inproceedings{reich_diffjpeg,
  author    = {Reich, Christoph and Debnath, Biplob and Patel, Deep and Chakradhar, Srimat},
  title     = {Differentiable {JPEG}: The Devil is in the Details},
  booktitle = {IEEE/CVF Winter Conference on Applications of Computer Vision (WACV)},
  year      = {2024}
}

\end{document}